\documentclass[conference]{IEEEtran}
\IEEEoverridecommandlockouts
\usepackage{cite}
\usepackage{amsmath,amssymb,amsfonts}
\usepackage{algorithmic}
\usepackage{graphicx}
\usepackage{textcomp}
\usepackage{xcolor}
\usepackage{booktabs}
\usepackage{caption} 
\usepackage{hyperref} 
\usepackage{graphicx}
\usepackage{float} 
\hypersetup{
  colorlinks=true,
  linkcolor=black,
  citecolor=black,
  filecolor=magenta,
  urlcolor=black,
  pdfborder={0 0 0},
}
\def\BibTeX{{\rm B\kern-.05em{\sc i\kern-.025em b}\kern-.08em
    T\kern-.1667em\lower.7ex\hbox{E}\kern-.125emX}}
\begin{document}

\title{LINK-KG: LLM-Driven Coreference-Resolved Knowledge Graphs for Human Smuggling Networks
}

\author{\IEEEauthorblockN{Dipak Meher}
\IEEEauthorblockA{\textit{Department of Computer Science} \\
\textit{George Mason University}\\
Fairfax VA, USA \\
dmeher@gmu.edu}
\and
\IEEEauthorblockN{Carlotta Domeniconi}
\IEEEauthorblockA{\textit{Department of Computer Science} \\
\textit{George Mason University}\\
Fairfax VA, USA \\
cdomenic@gmu.edu}
\and
\IEEEauthorblockN{Guadalupe Correa-Cabrera}
\IEEEauthorblockA{\textit{Schar School of Policy and Government} \\
\textit{George Mason University}\\
Arlington VA, USA \\
gcorreac@gmu.edu}
}

\maketitle

\begin{abstract}
Human smuggling networks are complex and constantly evolving, making them difficult to analyze comprehensively. Legal case documents offer rich factual and procedural insights into these networks but are often long, unstructured, and filled with ambiguous or shifting references, posing significant challenges for automated knowledge graph (KG) construction. Existing methods either overlook coreference resolution or fail to scale beyond short text spans, leading to fragmented graphs and inconsistent entity linking. We propose LINK-KG, a modular framework that integrates a three-stage, LLM-guided coreference resolution pipeline with downstream KG extraction. At the core of our approach is a type-specific Prompt Cache, which consistently tracks and resolves references across document chunks, enabling clean and disambiguated narratives for structured knowledge graph construction from both short and long legal texts. LINK-KG reduces average node duplication by 45.21\% and noisy nodes by 32.22\% compared to baseline methods, resulting in cleaner and more coherent graph structures. These improvements establish LINK-KG as a strong foundation for analyzing complex criminal networks\footnotemark. 
\end{abstract}
\footnotetext{Note: All example names in this paper are anonymized or fictitious to protect privacy. Full proper names and company names are replaced with initials followed by a period (e.g., ``L.R.C.'').}


\section{Introduction}

Human smuggling networks represent highly adaptive and organized systems involving a web of actors, routes, vehicles, and intermediaries, often operating under the radar of restrictive immigration policies \cite{carrasco2025scapegoating}. These networks exploit legal loopholes, adjust swiftly to enforcement changes, and frequently intersect with transnational criminal organizations. Effectively analyzing their structure and behavior is critical for informing policy, enhancing security, and preventing exploitation. However, much of the actionable insight remains embedded in lengthy, unstructured legal documents, such as court rulings, field reports, and case transcripts, making automated analysis both essential and challenging.

Despite growing interest from legal and social science communities, computational approaches for analyzing these documents remain underdeveloped. Entity references in unstructured legal text are frequently inconsistent—appearing as aliases, abbreviations, or role-based titles (e.g., ``Officer Ross'' vs. ``Defendant Ross'') - which complicates coreference resolution, entity normalization, and downstream tasks like entity extraction and knowledge graph construction.

Prior work has shown the utility of knowledge graphs in legal investigations, typically using rule-based or regex-driven methods to extract information from case documents~\cite{mazepa2022relationships}. While effective in controlled settings, these methods depend on fixed templates and fail to handle aliasing or entity variation, limiting their applicability to complex narratives.

Recent LLM-based systems improve graph construction through modular, prompt-driven pipelines with type-aware coreference resolution and structured entity-relation extraction~\cite{meher2025llm}. However, they struggle with long documents due to the loss-in-the-middle problem, where mid-text content receives less attention due to positional encoding limits. More importantly, they often lack intermediate structured outputs to trace aliasing decisions, crucial in legal cases involving role shifts (e.g., a smuggler later referred to as a driver), ambiguous aliases (e.g., “the agent” used for multiple individuals), or plural references (e.g., “the agents” denoting multiple people). These challenges are especially acute in human smuggling cases, where references frequently shift and overlap.

Recent work such as \textsc{LLMLINK}~\cite{zhu2025llmlink} leverages dual LLMs and reference tracking to improve coreference resolution in long texts. While effective on structured, homogeneous datasets like book chapters, it struggles with complex texts involving multi-role entities, shifting aliases, and plural references. To address these challenges, we propose \textsc{LINK-KG}, a modular LLM-guided framework extending \textsc{LLMLINK} with type-specific entity tracking. Using tailored prompts for plural mentions and role shifts, \textsc{LINK-KG} generates intermediate representations suitable for coherent, disambiguated knowledge graph construction across short and long documents.

\textsc{Link-KG} comprises two core components:
(1) A two-stage, prompt-based coreference module that leverages instruction-tuned LLMs to capture both short- and long-range references. It constructs a type-specific Prompt Cache linking noun phrases to canonical names with auxiliary descriptions, enabling structured resolution of semantically or contextually equivalent mentions.
(2) A knowledge graph construction module that employs structured prompts with domain-specific filtering, sequential entity-type extraction, and explicit type definitions. This design reduces attention spread and misclassification, enabling accurate extraction of entities and relationships from legal texts.

 Our method handles short and long legal documents, improves coreference consistency, and enables interpretable knowledge graph construction. The key contributions are:\footnote{Our code is available at: \url{https://github.com/dipakmeher/LinkKG-HS}}

\begin{itemize}
   \item We introduce \textsc{Link-KG}, a modular LLM-based framework that combines type-specific coreference resolution, supported by a persistent Prompt Cache, with domain-specific prompting for accurate entity and relationship extraction from complex human smuggling cases.

    \item We develop a prompt-guided resolution module that builds a type-specific Prompt Cache, mapping noun phrases to canonical forms with auxiliary descriptions. This enables resolution of plural aliases, role shifts, and context-dependent references across local and global spans, addressing key challenges in legal narratives.

    \item We evaluate \textsc{Link-KG} on U.S. federal and state court documents related to human smuggling, achieving a 45.21\% reduction in node duplication and a 32.22\% drop in noisy nodes compared to existing baselines.
\end{itemize}

\section{Related Work}

Knowledge graphs (KGs) are widely used to transform unstructured text into structured formats that support reasoning, analysis, and retrieval~\cite{edge2024local}. Their applications span multiple domains, including education~\cite{chen2018knowedu}, life sciences~\cite{callahan2024open}, and legal investigations~\cite{mazepa2022relationships}. However, without careful design, KGs often suffer from node duplication and fragmented structures that hinder downstream utility~\cite{huaman2020duplication}.

\begin{figure*}[htbp]
    \centering
    \includegraphics[width=\textwidth]{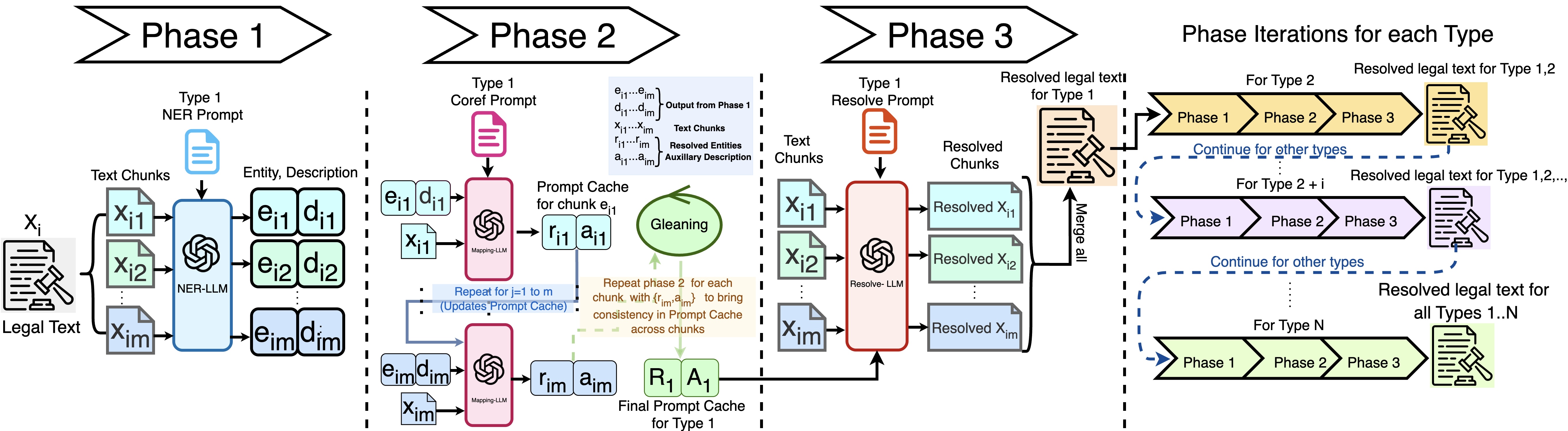}
    \caption{Overview of the LINK-KG framework. The pipeline operates in three main phases for each entity type. In Phase 1, the legal text is split into chunks and passed through a NER-LLM to extract type-specific entities and their descriptions. In Phase 2, a Mapping-LLM takes these outputs along with previously cached mappings to iteratively update a prompt cache for each chunk, ensuring consistency through a “gleaning” step across all chunks. In Phase 3, the Resolve-LLM uses the final prompt cache to perform coreference resolution and generate resolved legal chunks, which are then merged to produce the resolved output for that entity type. This three-phase process is repeated sequentially for each entity type (Type 1 to Type N) to obtain the final coreference-resolved legal text.
}
    \label{fig:linkkg_pipeline}
\end{figure*}

\subsection{Knowledge Graph Construction}

Traditional knowledge graph construction extracts entities and relationships from text using rule-based~\cite{sun2018overview}, dictionary-driven, and statistical or machine learning methods~\cite{liu2011recognizing}. Relation extraction approaches include syntactic, semantic~\cite{kambhatla2004combining}, ontology-based~\cite{liu2007implementation}, and semi-supervised frameworks~\cite{carlson2010coupled}. Although effective in narrow domains, these techniques depend heavily on handcrafted rules and surface-level cues, limiting adaptability to complex or evolving contexts.

Recent LLM-based frameworks~\cite{kommineni2024human,khorashadizadeh2023exploring} have shown promise for prompt-driven knowledge graph construction by combining tasks such as information extraction, schema definition, and ontology-guided triple generation. While effective at generating relevant triples, these systems often assume clean, unambiguous input and lack mechanisms for handling reference ambiguity, aliasing, role-shifting, and coreference resolution. These capabilities are critical for constructing robust and coherent knowledge graphs from complex texts.

To address such challenges, coreference resolution is essential for maintaining graph coherence. Wang et al.~\cite{wang2020coreference} showed that even sparse pronoun references can significantly affect knowledge graph completeness in the educational domain, highlighting the need to integrate coreference mechanisms into the construction pipeline when dealing with implicit or inconsistent entity mentions.

Prior work has applied knowledge graph construction to criminal domains. Mazepa et al.\cite{mazepa2022relationships} built a rule-based graph for homicide investigations using CoreNLP, while Shi et al.\cite{shi2022knowledge} used regular expressions to extract entities and relations for job-related crime indictments in Neo4j. Though effective in structured scenarios, these systems use static templates, lack coreference resolution or modular prompting, limiting scalability in handling ambiguous, multi-entity legal narratives.

A recent LLM-based framework, CORE-KG~\cite{meher2025llm}, enables automatic knowledge graph construction from legal case documents using type-specific coreference resolution and structured prompting. However, it struggles with long documents due to the loss-in-the-middle issue in LLMs, resulting in fragmented and inconsistent graphs. \textsc{Link-KG} addresses these limitations by introducing a persistent Prompt Cache that maps noun phrases to their canonical forms and uses this mapping to resolve coreferences chunk by chunk, enabling coherent and accurate graph construction across both short and long texts.

\subsection{Coreference Resolution}
Several methods have been proposed to mitigate duplicate nodes and fragmented structures through coreference resolution~\cite{liu2023brief}. Pogorilyy et al.~\cite{pogorilyy2019coreference} and Wu et al.~\cite{wu2017deep} each proposed coreference resolution methods based on convolutional neural networks (CNNs), demonstrating improvements in modeling semantic and syntactic patterns in text.

Recent advances in large language models have boosted coreference resolution, particularly in low-resource settings without annotated data. Prompt-based methods perform well using few-shot, zero-shot, or chain-of-thought strategies~\cite{das2024co}, but most work targets single entity types—typically human mentions—or general narratives without explicit type modeling~\cite{tran2025coreference}. While effective for focused attention, such setups fail to capture the complexity of domain-specific texts, where diverse entities are inconsistently referenced across roles. Thus, LLMs’ ability to generalize across types and resolve role-shifting references remains underexplored.

Recent advances in long-form coreference resolution, like \textsc{LLMLINK}~\cite{zhu2025llmlink}, adopt memory mechanisms, a prompt cache to incrementally track alias-to-canonical mappings. However, they face challenges such as inconsistent handling of plural mentions, often resolving them to singular entities, and context overflow from an expanding prompt cache. These issues primarily result from type-agnostic behavior.

Some progress has been made in coreference resolution for legal texts. Jia et al.\cite{ji2020deep} proposed a neural model combining ELMo, BiLSTM, and Graph Convolutional Networks to resolve speaker-based coreference in court records. More recently, CORE-KG\cite{meher2025llm} employs instruction-tuned LLMs with type-aware prompting to handle multi-entity legal documents. While effective for shorter documents, such approaches are constrained by the loss-in-the-middle problem and struggle to maintain consistent reference chains across lengthy narratives.

To address these limitations, \textsc{LINK-KG} adopts a memory-based coreference resolution strategy inspired by \textsc{LLMLINK}~\cite{zhu2025llmlink}. Specifically, we use type-specific prompt caches that store alias-to-canonical mappings separately for each entity type (e.g., person, location), allowing precise disambiguation within types while keeping token usage low in the LLM’s context window. We further design plural-aware prompting strategies and context-sensitive alias addition to accurately resolve plural references and prevent redundant mappings of previously resolved entities.

\section{Method}

Our proposed pipeline enables modular knowledge graph (KG) construction from narrative-rich documents. While broadly applicable across domains, we focus on legal case files, particularly those involving human smuggling networks. The pipeline is designed to identify actors, events, and contextual entities, and extracts seven key entity types: \textit{Person}, \textit{Location}, \textit{Organization}, \textit{Route}, \textit{Means of Transportation}, \textit{Means of Communication}, and \textit{Smuggled Items}. It comprises two major components: a prompt-based coreference resolution module and a KG construction module. The coreference module incrementally constructs a type-specific prompt cache to resolve references across document chunks. The KG module then employs a tailored prompt to extract entities and relations, which are subsequently assembled into a structured knowledge graph using the GraphRAG framework.



\subsection{Coreference Resolution}


We propose a prompt-based, three-stage pipeline for coreference resolution that operates effectively on both short and long legal documents, as illustrated in Figure~\ref{fig:linkkg_pipeline}.

\subsubsection{Stage 1: NER using LLM}
Recent work shows that large language models (LLMs), when guided by structured prompts, outperform traditional systems in named entity recognition (NER) tasks~\cite{zhou2023universalner}. Leveraging this, we use an instruction-tuned LLM—referred to as NER-LLM—to extract proper nouns, noun phrases, and their descriptions from each text chunk.

\paragraph{Stage 1 Prompt Design}  
We use structured, instruction-based prompts specific to each entity type (e.g., \texttt{Person}, \texttt{Location}, \texttt{Organization}). Each prompt begins with a goal definition, followed by a rule-based sequence guiding the LLM through consistent extraction steps: identifying \texttt{PROPER\_NOUN} mentions (explicit names or titles), extracting \texttt{NOUN\_PHRASE} mentions (descriptive or role-based references), filtering non-target entities, generating brief descriptions for each proper noun, and formatting the output as valid JSON. The final output includes two components: \texttt{ENTITIES}, listing all \texttt{PROPER\_NOUN} and \texttt{NOUN\_PHRASE} mentions, and \texttt{PROPER\_NOUN\_DESCRIPTION}, mapping each \texttt{PROPER\_NOUN} to a concise role description. This structured format ensures precision and smooth integration with the coreference resolution stage.

\paragraph{NER Stage Formulation}
Following the notation of \textsc{LLMLINK}~\cite{zhu2025llmlink}, let $X_i$ be a legal document divided into chunks $\{x_{i1}, x_{i2}, \ldots, x_{im}\}$, where $x_{ij}$ is the $j$-th chunk and $m$ is the total number of chunks. For each chunk, the NER-LLM extracts entities and corresponding descriptions, denoted by $\{e_{ij}, d_{ij}\}$. Each $e_{ij}$ contains pairs of proper nouns and noun phrases: $e_{ij} = \{(p, n) \mid p \in \text{ProperNoun}_{ij},\ n \in \text{NounPhrase}_{ij}\}$. Let $p\text{-}ner$ denote the prompt for the \texttt{Person} entity type. Then, Stage~1 is defined as:
\[
\{e_{ij}, d_{ij}\} = \text{NER-LLM}(x_{ij}, p\text{-}ner)
\]
Here, $d_{ij}$ maps each proper noun to a brief role-based description. The process is performed independently for each chunk, and results are passed to the next stage.

\subsubsection{Stage 2: Constructing Prompt Cache using LLM}

In Stage~2, we use a prompt-guided LLM, Mapping-LLM, to construct a prompt cache that maps noun phrases (e.g., aliases, roles, abbreviations) to their canonical noun forms. The cache also stores auxiliary descriptions from Stage~1 to support disambiguation during coreference resolution.

\paragraph{Stage 2 Prompt Design} 
We use Mapping-LLM to construct a prompt cache that links noun phrases to canonical names for each entity type (e.g., \texttt{Person}, \texttt{Location}). The prompt begins by specifying the target entity type and guiding the model through a structured resolution process. Unambiguous aliases are linked to existing canonical names, while unmatched ones generate new entries. When aliases refer to different entities in different contexts, the prompt instructs the model to use minimal surrounding context for disambiguation. Plural mentions are resolved by linking to multiple canonical names when specific entities are implied or mapped to \texttt{null} if ambiguous. Vague or unresolvable aliases are also assigned \texttt{null}. Auxiliary descriptions are updated only with explicit, non-speculative details. The output is a strict JSON object with two fields: \texttt{RESOLVED\_ENTITIES} for alias-to-canonical mappings and \texttt{AUXILIARY\_DESCRIPTIONS} for concise role-based summaries.

One of the key novelties of our work is the design of a prompt that robustly handles shifting references, plural noun phrases, and ambiguous aliases, which are challenges commonly encountered in legal narratives.

Shifting references arise when the same role label refers to different individuals. For example, “the driver” may denote a detained defendant in one sentence and a patrol agent in another. Vague aliases like “the agent” can similarly point to different people across document chunks. Without disambiguation, these may be wrongly linked to the same canonical entity, resulting in false coreference and misleading graph edges. To prevent this, our prompt enforces context-aware alias generation using surrounding clues, such as “the driver in a truck” vs. “the driver in a patrol car”, or “the agent from Checkpoint A” vs. “the agent from Checkpoint B”. Auxiliary descriptions further improve accuracy by encoding key contextual traits of each canonical entity.

Plural noun phrases like “the defendants” or “the passengers” are often mistakenly linked to a single nearby person. To avoid this, our prompt enforces explicit grounding: resolution is allowed only if all individuals are named in the text. For example, “the defendants” maps to “NameX, NameY, and NameZ” if all three are mentioned; otherwise, it resolves to null. This prevents speculative coreference and ensures precise alignment between plural mentions and canonical entities, crucial for accurate resolution knowledge graph construction.

\paragraph{Stage 2 Formulation}
The outputs from Stage~1, denoted as $\{e_{ij}, d_{ij}\}$ for chunk $x_{ij}$ of document $i$, are passed to the Mapping-LLM along with the type-specific prompt $p\text{-}map$ and the current alias-to-canonical mappings and auxiliary descriptions, denoted by $\{r_{ij}, a_{ij}\}$.

The model outputs an updated set of mappings and descriptions:
\[
\{r_{ij+1}, a_{ij+1}\} = \text{Mapping-LLM}(e_{ij}, d_{ij}, r_{ij}, a_{ij}, p\text{-}map)
\]
Here, $r_{ij+1}$ and $a_{ij+1}$ represent the prompt cache updated after processing chunk $x_{ij}$. This process is repeated for each chunk, accumulating a global mapping over the document. 

Let $\{R, A\}$ denote the resolved entity dictionary and auxiliary descriptions after the first pass. To improve consistency, an optional second pass (\emph{gleaning}) reruns Stage~2 with $\{R, A\}$ as input, allowing the model to revise earlier ambiguous or incomplete mappings using canonical names and context observed later in the text, yielding refined outputs:
\[
\{R, A\} \leftarrow \text{Mapping-LLM}(e_{ij}, d_{ij}, R, A, p\text{-}map)
\]
This gleaning step ensures that alias-to-canonical mappings are globally coherent and that auxiliary descriptions reflect the most complete and accurate context for each canonical entity.

\subsubsection{Stage 3: Coreference Resolution}

In the final stage, we perform chunk-wise coreference resolution using an LLM (Resolve-LLM). Each chunk is processed with the Stage 2 prompt cache, which contains alias-to-canonical mappings and auxiliary descriptions, along with a structured prompt guiding resolution. Naive alias substitution often fails when the same alias refers to different entities across contexts (e.g., “the agent” referring to “Agent Neil Martin” at one checkpoint and “Agent Jack Allen” at another). Our method resolves such ambiguities by leveraging both local context and auxiliary descriptions to assign the correct canonical name.

\paragraph{Stage 3 Prompt Design}  
Resolve-LLM uses a structured prompt to rewrite each chunk by replacing aliases with their canonical names from \texttt{RESOLVED\_ENTITIES}. It first detects all aliases, accounting for variations in punctuation, quotation, and casing, then performs exact substitutions while preserving sentence flow, grammar, and legal tone. If an alias maps to multiple entities, the full multi-name string is inserted. Auxiliary descriptions from Stage 2 aid internal disambiguation but are excluded from the output. The prompt strictly prohibits hallucination or inference beyond the mappings. The result is a coreference-resolved paragraph ready for knowledge graph construction.

\paragraph{Stage 3 Formulation}
Let $x_{ij}$ denote the $j$-th chunk of document $i$, and $\{R, A\}$ be the globally accumulated alias-to-canonical mappings and auxiliary descriptions from Stage~2. Let $p\text{-}resolve$ be the prompt used for coreference resolution of a specific entity type (e.g., \texttt{Person}). The resolved output is given by:
\[
x^{\text{resolved}}_{ij} = \text{Resolve-LLM}(x_{ij}, R, A, p\text{-}resolve)
\]

Here, \texttt{Resolve-LLM} applies $p\text{-}resolve$ to substitute aliases in $x_{ij}$ with canonical names from $R$, using $A$ for context-aware disambiguation. This chunk-wise rewriting ensures legally consistent text suitable for downstream structured knowledge graph construction. Resolved chunks are then merged to form final output for the corresponding entity type.

\begin{figure}[t]  
    \centering
    \includegraphics[width=\columnwidth]{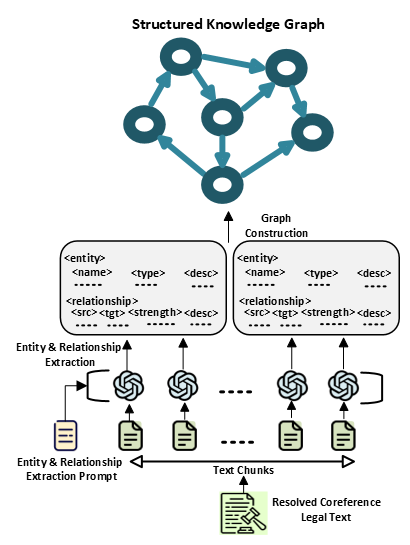}
    \caption{Overview of Knowledge Graph Construction Module. The resolved coreference legal document is split into chunks. Each chunk is paired with a prompt and sent to an LLM to extract entities and relationships. The extracted information is then combined to build a structured knowledge graph.}
    \label{fig:kgconst}
\end{figure}

\subsection{Entity-Relationship Extraction and KG Construction}

Figure~\ref{fig:kgconst} shows the Knowledge Graph Construction (KGC) pipeline. Coreference-resolved text is split into overlapping 300-token chunks and processed by GraphRAG~\cite{edge2024local} using a domain-optimized extraction prompt. The LLM generates entity–relationship triples, which are aggregated and post-processed by merging entities with exact string and type matches. The final graph is assembled using \texttt{NetworkX}.


\paragraph{Prompt Design for Entity and Relationship Extraction}
GraphRAG’s KGC component uses a unified prompt to extract all seven entity types and their relationships in a single pass. To ensure both breadth and precision, the prompt is modular and step-wise, guiding the LLM to generate structured triples with minimal noise and high relevance.

First, it defines the task and assigns the LLM the role of an expert extractor of factual entities and relationships in the context of human smuggling. Second, it enumerates the seven target entity types with brief, domain-specific definitions to ensure type-aware extraction. Third, it outlines the procedure for entity extraction—requiring the model to identify entity names, assign types, and generate concise descriptions.

Fourth, the prompt defines a relationship extraction protocol that instructs the model to identify explicit links between entities, assign strength scores based on textual cues, and output structured triples. Fifth, it incorporates domain-specific filtering rules to remove legal or government entities that do not support smuggling network analysis. Finally, few-shot examples demonstrate accurate extraction of both entities and relationships.

Below, we outline additional prompt strategies employed to further enhance extraction quality.

\paragraph{Sequential Entity Extraction to Mitigate Attention Spread}
GraphRAG’s joint extraction prompt often leads to attention diffusion, causing missed entities or incorrect typing in dense legal texts~\cite{abdelnabi2025get}. To mitigate this, we enforce a fixed extraction order where entities are extracted by type (e.g., \textit{Person}, \textit{Location}, \textit{Routes}, etc.) before relationship extraction. This reduces inter-type attention interference, enhancing recall, typing accuracy, and graph structure.

\paragraph{Filtering High-Frequency Irrelevant Entities}
Legal texts often contain high-frequency but irrelevant entities—such as courts, juries, and appeals—typically tagged as \textit{Organization}. Including them can clutter the graph with unrelated nodes. To prevent this, the prompt includes a filtering step where the LLM removes government-related entities using predefined rules. This in-prompt filtering reduces post-processing and produces cleaner, more focused graphs.

\paragraph{Entity Type Definitions to Reduce Overgeneralization Bias}
LLMs often misclassify entities by relying on patterns from pretraining rather than the local context. For example, they may incorrectly label event names as \textit{Location} because such terms frequently appear near geographic names in training data~\cite{peters2025generalization}. This issue is common in legal texts where overlapping terminology is frequent. To address this, we include clear definitions and examples for all seven entity types (such as \textit{Person}, \textit{Route}, \textit{Organization}) directly in the prompt. These definitions help the model focus on the document context, improving classification accuracy and consistency.

\setlength{\tabcolsep}{3pt}
\begin{table*}[t]
\centering
\scriptsize
\begin{tabular}{@{}lc|
ccc|ccc|ccc|
ccc|ccc|ccc@{}}
\toprule
Case & 
& \multicolumn{9}{c|}{\textbf{Node Duplication}} 
& \multicolumn{9}{c}{\textbf{Noisy Nodes}} \\
\cmidrule(lr){3-11} \cmidrule(l){12-20}
& 
& \multicolumn{3}{c|}{GraphRAG} 
& \multicolumn{3}{c|}{CORE-KG} 
& \multicolumn{3}{c|}{LINK-KG} 
& \multicolumn{3}{c|}{GraphRAG} 
& \multicolumn{3}{c|}{CORE-KG} 
& \multicolumn{3}{c}{LINK-KG} \\
\cmidrule(lr){3-5} \cmidrule(lr){6-8} \cmidrule(lr){9-11}
\cmidrule(lr){12-14} \cmidrule(lr){15-17} \cmidrule(l){18-20}
& 
& Total & Dup. Ent & \% 
& Total & Dup. Ent & \% 
& Total & Dup. Ent & \% 
& Total & Noisy Ent & \% 
& Total & Noisy Ent & \% 
& Total & Noisy Ent & \% \\
\midrule
Case 01 & & 94 & 32 & 34.04 & 51 & 13 & 25.49 & 46 & 7 & \textbf{15.22} & 94 & 26 & 27.66 & 51 & 5 & 9.80 & 46 & 3 & \textbf{6.52} \\
Case 02 & & 86 & 24 & 27.91 & 42 & 5 & 11.90 & 52 & 7 & \textbf{13.46} & 86 & 28 & 32.56 & 42 & 0 & \textbf{0.00} & 52 & 2 & 3.85 \\
Case 03 & & 60 & 19 & 31.67 & 22 & 5 & 22.73 & 22 & 3 & \textbf{13.64} & 60 & 17 & 28.33 & 22 & 5 & \textbf{22.73} & 22 & 6 & 27.27 \\
Case 04 & & 75 & 15 & 20.00 & 34 & 5 & 14.71 & 34 & 2 & \textbf{5.88}  & 75 & 19 & 25.33 & 34 & 7 & 20.59 & 34 & 5 & \textbf{14.71} \\
Case 05 & & 49 & 11 & 22.45 & 21 & 3 & 14.29 & 21 & 3 & \textbf{14.29} & 49 & 9  & 18.37 & 21 & 3 & \textbf{14.29} & 21 & 4 & 19.05 \\
Case 06 & & 68 & 20 & 29.41 & 57 & 15 & 26.32 & 32 & 2 & \textbf{6.25}  & 68 & 10 & 14.71 & 57 & 5 & 8.77  & 32 & 2 & \textbf{6.25} \\
Case 07 & & 55 & 13 & 23.64 & 28 & 1 & \textbf{3.57}  & 36 & 2 & 5.56  & 55 & 10 & 18.18 & 28 & 3 & 10.71 & 36 & 3 & \textbf{8.33} \\
\midrule
\multicolumn{2}{r|}{\textbf{Average}} 
& 69.57 & 19.14 & 27.02 
& 36.43 & 6.71 & 17.00 
& 34.71 & 3.71 & \textbf{10.61} 
& 69.57 & 17.00 & 23.59 
& 36.43 & 4.00 & 12.41 
& 34.71 & 3.57 & \textbf{12.28} \\
\bottomrule
\end{tabular}

\vspace{0.5em}
\caption{
Comparison of node duplication and legal noise  across different models for short legal case documents. 
{\textit{Total} is the total number of extracted entities; \textit{Dup. Ent}   is
the count  of duplicate entities. \textit{Noisy Ent}  is the count  of noisy entities. The percentages indicate the proportion of duplicate or noisy entities relative to the total extracted entities in the respective method.
}}
\label{tab:shorter_stats_entities_extracted}
\end{table*}

\setlength{\tabcolsep}{3pt}
\begin{table*}[t]
\centering
\scriptsize
\begin{tabular}{@{}lc|
ccc|ccc|ccc|
ccc|ccc|ccc@{}}
\toprule
Case & 
& \multicolumn{9}{c|}{\textbf{Node Duplication}} 
& \multicolumn{9}{c}{\textbf{Noisy Nodes}} \\
\cmidrule(lr){3-11} \cmidrule(l){12-20}
& 
& \multicolumn{3}{c|}{GraphRAG} 
& \multicolumn{3}{c|}{CORE-KG} 
& \multicolumn{3}{c|}{LINK-KG} 
& \multicolumn{3}{c|}{GraphRAG} 
& \multicolumn{3}{c|}{CORE-KG} 
& \multicolumn{3}{c}{LINK-KG} \\
\cmidrule(lr){3-5} \cmidrule(lr){6-8} \cmidrule(lr){9-11}
\cmidrule(lr){12-14} \cmidrule(lr){15-17} \cmidrule(l){18-20}
& 
& Total & Dup. Ent & \% 
& Total & Dup. Ent & \% 
& Total & Dup. Ent & \% 
& Total & Noisy Ent & \% 
& Total & Noisy Ent & \% 
& Total & Noisy Ent & \% \\
\midrule
Case 08 & & 83 & 20 & 24.10 & 46 & 7 & 15.22 & 40 & 5 & \textbf{12.50} & 83 & 54 & 65.06 & 46 & 20 & 43.48 & 40 & 10 & \textbf{25.00} \\
Case 09 & & 131 & 50 & 38.17 & 76 & 29 & 38.16 & 80 & 15 & \textbf{18.75} & 131 & 33 & 25.19 & 76 & 17 & 22.37 & 80 & 16 & \textbf{20.00} \\
Case 10 & & 104 & 37 & 35.58 & 45 & 11 & 24.44 & 50 & 10 & \textbf{20.00} & 104 & 26 & 25.00 & 45 & 6 & 13.33 & 50 & 6 & \textbf{12.00} \\
Case 11 & & 149 & 56 & 37.58 & 37 & 8 & 21.62 & 64 & 12 & \textbf{18.75} & 149 & 58 & 38.93 & 37 & 6 & 16.22 & 64 & 4 & \textbf{6.25} \\
Case 12 & & 171 & 65 & 38.01 & 54 & 18 & 33.33 & 45 & 8 & \textbf{17.78} & 171 & 125 & 73.10 & 54 & 23 & 42.59 & 45 & 17 & \textbf{37.78} \\
Case 13 & & 183 & 59 & 32.24 & 67 & 11 & 16.42 & 63 & 9 & \textbf{14.29} & 183 & 104 & 56.83 & 67 & 17 & 25.37 & 63 & 15 & \textbf{23.81} \\
Case 14 & & 158 & 62 & 39.24 & 71 & 14 & 19.72 & 62 & 8 & \textbf{12.90} & 158 & 28 & 17.72 & 71 & 9 & 12.68 & 62 & 6 & \textbf{9.68} \\
Case 15 & & 99 & 28 & 28.28 & 48 & 13 & 27.08 & 46 & 8 & \textbf{17.39} & 99 & 38 & 38.38 & 48 & 8 & 16.67 & 46 & 6 & \textbf{13.04} \\
Case 16 & & 214 & 109 & 50.93 & 95 & 37 & 38.95 & 76 & 21 & \textbf{27.63} & 214 & 56 & 26.17 & 95 & 12 & 12.63 & 76 & 8 & \textbf{10.53} \\
\midrule
\multicolumn{2}{r|}{\textbf{Average}} 
& 143.56 & 54.00 & 36.01 
& 59.89 & 16.44 & 26.10 
& 58.44 & 10.67 & \textbf{17.78} 
& 143.56 & 58.00 & 40.71 
& 59.89 & 13.11 & 22.82 
& 58.44 & 9.78 & \textbf{17.57} \\
\bottomrule
\end{tabular}

\vspace{0.5em}
\caption{
Comparison of node duplication and legal noise  across different models for long legal case documents. 
{\textit{Total} is the total number of extracted entities; \textit{Dup. Ent}   is
the count  of duplicate entities. \textit{Noisy Ent}  is the count  of noisy entities. The percentages indicate the proportion of duplicate or noisy entities relative to the total extracted entities in the respective method.
}}
\label{tab:longer_stats_entities_extracted}
\end{table*}

\section{Experiments}

We aim to address the following research questions. (RQ1): Does the integration of cache-based coreference resolution reduce node duplication in knowledge graphs constructed from short and long human smuggling case documents?
(RQ2): Does LINK-KG
produce more relevant and concise knowledge graphs compared to a GraphRAG-based baseline?

\subsection{Dataset}
To evaluate our system, we use publicly available judicial cases on human smuggling accessed via Nexis Uni through George Mason University Library. These federal and state court documents, filed between 1994 and 2024, offer a diverse sample of smuggling activities across jurisdictions. Since legal texts often contain procedural content irrelevant to narrative analysis, we extract the ``Opinion’’ section, which provides detailed factual descriptions and actor interactions. All methods in our experiments use the same Opinion section, referred to as the input document.

In our experiments, we randomly selected 16 legal cases of varying lengths, including 7 shorter and 9 longer documents, retrieved from the Nexis Uni database using the query “human smuggling OR alien smuggling.” We define shorter documents as those with 2,500 words or fewer, and longer ones as those exceeding 2,500 words. This division allows us to evaluate model performance across different lengths, with results in Table~\ref{tab:shorter_stats_entities_extracted} for short cases and Table~\ref{tab:longer_stats_entities_extracted} for long cases.

\subsection{Implementation}

We use the open-source LLaMA 3.3 70B model for both coreference resolution and knowledge graph construction. To ensure reproducibility, we fix the temperature to 0. The model is served locally via the Ollama framework, enabling inference without commercial or closed-source APIs. All experiments run on an NVIDIA A100 GPU (80GB). Our implementation uses Python 3.12. For GraphRAG~\cite{edge2024local}, we adopt version 0.3.2 with default settings from the official GitHub. We configure 300-token chunks and specify the nomic-embed-text embedding model, though embeddings are unused in our pipeline.

\subsection{Baselines}

Given limited prior work on knowledge graph construction from legal texts, particularly for criminal network analysis like human smuggling, we compare our method against two baselines: GraphRAG and CORE-KG. GraphRAG adapts the prompt from~\cite{edge2024local} by incorporating our seven entity types and a few-shot example, while retaining its original format. CORE-KG~\cite{meher2025llm} is a domain-tuned pipeline combining prompt-based coreference resolution with graph construction for legal cases. As all three methods are LLM-based, they offer fair comparison points. For consistency, we apply a uniform chunking strategy of 300 tokens across all methods.

\subsection{Evaluation Measures}
Since no annotated ground truth exists for structured knowledge graphs in this domain, we adopt both a qualitative and quantitative evaluation framework. All methods are applied to the same 16 input documents. 
We evaluate the resulting graphs in terms of {\textit{duplicate node and noise rates}}.
We also qualitatively examine common misclassifications of entities (e.g., labeling ``The Safehouse'' as an organization instead of a location) and evaluate the impact of prompt-level refinements in reducing such errors.

    
    


We detect duplicate nodes in two stages. First, we use fuzzy string matching with the \texttt{partial\_ratio} function from the \texttt{RapidFuzz} library to compare intra-type entity pairs. Pairs with a similarity score $\geq$75\% are connected in an undirected similarity graph, and connected components are treated as duplicates. For example, \texttt{white pickup truck}, \texttt{stolen white pickup truck}, and \texttt{white older Ford pickup truck} are clustered together. In the second stage, a subject matter expert manually reviews the clusters to remove false positives and add missed duplicates.

The final node duplication count is computed as $\sum_{C_i} (|C_i| - 1)$, where $C_i$ denotes a cluster of mentions referring to the same entity. To account for differences in graph size, the node duplication count is normalized by dividing by the total number of nodes in the graph. 

To assess the quality and interpretability of the generated knowledge graphs, a domain expert validated extracted entities. This process involved identifying procedural or boilerplate terms (e.g., \texttt{Court}, \texttt{Judicial Proceedings}) that do not aid structural understanding of the smuggling network. The noise rate is defined as the percentage of such noisy nodes relative to total nodes in the graph.

\section{Results}

\subsection{RQ1: Impact of Cache-based Coreference Resolution on Node duplication}\label{sec:rq1_coref}

Table~\ref{tab:shorter_stats_entities_extracted} and Table~\ref{tab:longer_stats_entities_extracted} present average node duplication rates across 16 legal cases, categorized into short and long documents. \textsc{Link-KG} consistently outperforms both baselines, reducing duplication from 27.02\% (GraphRAG) and 17.00\% (CORE-KG) to 10.61\% for shorter cases, and from 36.01\% (GraphRAG) and 26.10\% (CORE-KG) to 17.78\% for longer cases. This corresponds to relative reduction of 60.72\% and 37.59\% for shorter cases, and 50.63\% and 31.86\% for longer ones. On average, \textsc{Link-KG} achieves 49.16\% reduction in node duplication across baselines for shorter documents and 41.25\% for longer ones. Averaging across both document categories, \textsc{Link-KG} achieves overall 45.21\% reduction in node duplication over baseline methods. These results demonstrate \textsc{Link-KG} effectively reduces duplication across legal documents of varying lengths, and narrative intricacy.

In a head-to-head comparison across cases, LINK-KG consistently reduces node duplication more effectively than both baselines. For shorter cases, CORE-KG slightly outperforms LINK-KG in Case 02 by a minimal margin. This is likely due to its lower entity count, which reduces the opportunity for duplication but may also limit recall.

For longer documents, LINK-KG clearly outperforms both GraphRAG and CORE-KG, highlighting the strength of its prompt-cached memory mechanism in maintaining consistent coreference resolution across chunks. GraphRAG performs no coreference resolution, while CORE-KG uses a single-pass LLM prompt over the full input to perform coreference resolution, which becomes unreliable for long documents due to context window constraints and the ‘loss-in-the-middle’ effect. In contrast, LINK-KG’s type-specific memory propagation enables robust alias resolution across segments, effectively reducing redundancy and improving graph coherence.



During our case-by-case analysis, we observed that LINK-KG is effective in handling both plural and singular noun phrases in coreference resolution. Specifically, LINK-KG resolves plural references to their corresponding named entities when all are explicitly mentioned; otherwise, it assigns \texttt{null} if no proper nouns are available.

In Case 20, two agents—S.P. and A.B.—were involved. LINK-KG correctly mapped plural mentions like “the agents” and “the border patrolmen” to them, and resolved aliases such as “the occupants” and “the passengers” to “M.D.J.G., L.R.C., and others,” where M.D.J.G. is a defendant and L.R.C. the driver. For vague phrases like “the male passengers” or “the female passengers,” LINK-KG returned null, as expected. In contrast, the baselines do not handle plural aliases separately, leaving such mentions unresolved and causing redundant nodes and graph fragmentation, ultimately reducing relationship coverage for key entities.

Moreover, LINK-KG  performs well on singular noun phrase resolution. In the same case, where two agents were later referred to individually by their last names, LINK-KG accurately resolved these mentions to their corresponding full names. While CORE-KG is able to handle such singular coreferences reasonably well in shorter cases, its performance degrades significantly in longer cases, often missing many references throughout the document. These examples highlight LINK-KG’s effectiveness in linking across varied noun phrases, which is reflected in its improved node duplication metrics.


LINK-KG demonstrates strong context-awareness by accurately distinguishing between multiple organizational entities of the same type. In Case 16, three courts are involved: the U.S. District Court, the U.S. Supreme Court, and the U.S. Court of Appeals. Although the term “court” is used generically throughout the text, LINK-KG correctly mapped it to the U.S. District Court based on context. Since the court entity was already present in the prompt cache, LINK-KG followed the prompt’s instruction to use contextual cues for disambiguation and added variants like “a supreme court” and “a court of appeal” as aliases of their canonical entities. Since the pipeline incorporates context from both auxiliary descriptions and the input text during the third stage of coreference resolution using the prompt cache, these variations were accurately resolved. 

\subsection{RQ2: Does LINK-KG produce more relevant and concise knowledge graphs compared to the baselines?}

To evaluate the structural improvements introduced by LINK-KG, we analyze its ability to reduce legal noise, irrelevant mentions such as “magistrate judge,” “senate,” or “sentencing hearing” that are common in legal texts but uninformative for human smuggling analysis. LINK-KG addresses this by unifying fragmented references through coreference resolution and incorporating in-prompt filtering during entity extraction. Coreference resolution is beneficial because, once references are resolved and unified, LLM attention becomes more focused and less prone to semantic drift, compared to scenarios where references remain scattered and unresolved.

Table~\ref{tab:shorter_stats_entities_extracted} and Table~\ref{tab:longer_stats_entities_extracted} present noise reduction statistics. On average, for shorter documents, \textsc{Link-KG} reduces the noise rate from 23.59\% (GraphRAG) and 12.41\% (CORE-KG) to 12.28\%, representing a relative improvement of 47.96\% over GraphRAG and 1.05\% over CORE-KG. The average reduction across baselines for shorter cases is 24.51\%. For longer cases, \textsc{Link-KG} demonstrates substantial gains, reducing the noise rate from 40.71\% (GraphRAG) and 22.82\% (CORE-KG) to 17.57\%. This corresponds to a relative reduction of 56.83\% compared to GraphRAG and 23.00\% compared to CORE-KG, with an average reduction of 39.91\% across baselines. Overall, averaging across both shorter and longer cases, \textsc{Link-KG} achieves a relative noise reduction of 32.22\% over baseline methods, underscoring its effectiveness in improving knowledge graph quality across document lengths.

Moreover, in per-case comparisons, LINK-KG consistently outperforms both baselines. For shorter documents, there are three instances—Case 2, Case 3, and Case 5—where CORE-KG shows slightly better performance, with a marginal reduction of 1–2 noisy nodes. These cases also exhibit a notably lower total entity count, suggesting that small absolute differences can result in amplified changes in percentage-based metrics. For longer documents, LINK-KG consistently demonstrates the benefits of prompt cache-based coreference and structured prompting. 

\section{Conclusion}

We introduced LINK-KG, a modular LLM-based framework for constructing interpretable knowledge graphs from complex legal texts, focusing on human smuggling cases. LINK-KG integrates a three-stage coreference pipeline and type-specific prompt cache to resolve long-range references, plural mentions, ambiguous aliases, and role-shifting entity mentions, which are challenges often missed by prior systems. Structured prompting and type-aware memory ensure consistent resolution across both short and long documents. Empirical results show that LINK-KG reduces node duplication by 45.21\% and noisy nodes by 32.22\%, resulting in cleaner graphs that support downstream tasks such as group detection, role attribution, temporal analysis, and event prediction.

\section*{Acknowledgment}
This material is based upon work supported by the U.S. Department of Homeland Security under Grant Award Number 17STCIN00001-08-00. 
Disclaimer: The views and conclusions contained in this document are those of the authors and should not be interpreted as necessarily representing the official policies, either expressed or implied, of the U.S. Department of Homeland Security. 
\bibliographystyle{IEEEtran}
\bibliography{bibliography}
\end{document}